\documentclass[sigconf]{acmart}

\usepackage{booktabs} 

\usepackage{subcaption}
\usepackage{soul}
\hypersetup{draft}
\usepackage{hyperref}

\acmDOI{10.1145/nnnnnnn.nnnnnnn} 
\acmISBN{978-x-xxxx-xxxx-x/YY/MM} 
\acmConference[GECCO '22]{The Genetic and Evolutionary Computation Conference 2022}{July 9--13, 2022}{Boston, USA}
\acmYear{2022}
\copyrightyear{2022}

\begin{document}

\title{Reproducibility and Baseline Reporting for Dynamic Multi-objective Benchmark Problems}
\author{Daniel Herring}
\affiliation{\institution{University of Melbourne, Australia}\country{}} 
\affiliation{\institution{University of Birmingham, UK}\country{}} \email{dgh756@student.bham.ac.uk}
\author{Michael Kirley}
\affiliation{\institution{University of Melbourne, Australia}\country{}} 
\author{Xin Yao}
\affiliation{\institution{University of Birmingham, UK}\country{}}
\date{January 2022}

\newcommand\mk{\textcolor{purple}}

\renewcommand{\shortauthors}{D. Herring et al.}
\begin{abstract}

Dynamic multi-objective optimization problems (DMOPs) are widely accepted to be more challenging than stationary problems due to the time-dependent nature of the objective functions and/or constraints. Evaluation of purpose-built algorithms for DMOPs is often performed on narrow selections of dynamic instances with differing change magnitude and frequency or a limited selection of problems. In this paper, we focus on the reproducibility of simulation experiments for parameters of DMOPs. Our framework is based on an extension of PlatEMO, allowing for the reproduction of results and performance measurements across a range of dynamic settings and problems. A baseline schema for dynamic algorithm evaluation is introduced, which provides a mechanism to interrogate performance and optimization behaviours of well-known evolutionary algorithms that were not designed specifically for DMOPs. Importantly, by determining the maximum capability of non-dynamic multi-objective evolutionary algorithms, we can establish the minimum capability required of purpose-built dynamic algorithms to be useful. The simplest modifications to manage dynamic changes introduce diversity. Allowing non-dynamic algorithms to incorporate mutated/random solutions after change events determines the improvement possible with minor algorithm modifications. Future expansion to include current dynamic algorithms will enable reproduction of their results and verification of their abilities and performance across DMOP benchmark space.

\end{abstract}





\keywords{dynamic multi-objective optimization, dynamic optimization, dynamic benchmarks, dynamic parameters, dynamic frequency and severity}

\maketitle

\section{Introduction}

Optimisation of competing goals is the basis of multi-objective optimization. This is made more difficult when a time-dependent component is introduced, creating a dynamic multi-objective optimization problem (DMOP). There has been considerable work in the past two decades focused on the design of benchmark problems, the development of algorithms, performance measurements and visualization, however, there is limited understanding of the parameters controlling the dynamics. Existing works promote the use of a diverse problem set but carry out experiments for a limited range of the parameters controlling dynamic changes - restricting the true scope of any conclusions made.


Some of the principal benchmark dynamic multi-objective optimization problems (DMOPs) 
were defined nearly two decades ago by Farina et al. \cite{Farina2003}. Since then, a plethora of benchmark problems have been proposed  \cite{Jiang2017,Helbig2013a,Ruan2021,Jiang2019,HZhang2021,QZhang2021,Tang2007,Huang2011,Yazdani2020,Chen2017,Gee2017}, allowing for the testing of different characteristics of real world systems in a controllable environment. Dynamic Optimization Problem (DOP) generators such as Moving Peaks \cite{Branke1999,Yazdani2020} and the Dynamic XOR \cite{Yang2003} and others are well-known single-objective environments in which the difficulty of a problem instance can be controlled by increasing the number of peaks or bits respectively. Comprehensive surveys of methods, measurements and problems are given in \cite{Nguyen2012,Branke2002,Branke2003,Azzouz2016,Raquel2013,Helbig2013a,Helbig2013b}.


In existing dynamic multi-objective benchmarks, defining the `difficulty' of a particular problem or instance can be linked to characteristics of the dynamics in the problem. Farina et al. \cite{Farina2003} defined a four-type system according to whether the dynamics alter the: (I) Pareto-Optimal Set (POS), (III) Pareto-Optimal Front (POF), both (II) or neither (IV). Meanwhile, Nguyen et al. \cite{Nguyen2012} suggests a broader framework for DOPs to classify problems based on whether they have the following properties: Time-linkage; predictability of changes; visibility of dynamics to the optimization algorithm; constraints or dynamic constraints; single or multiple objectives; periodical, recurrent or cyclic dynamics and type and factor of dynamic changes (e.g. objective functions, decision variable domain or number). Azzouz et al. \cite{Azzouz2017} suggests a streamlined version of this for DMOPs that, in addition to the POS/POF observations, problems can be grouped based on the frequency, severity or predictability of changes.

Nevertheless, despite these frameworks for classifying the proposed benchmarks, there is little associated literature outlining the consistency of parameter settings and reproducibility of results on DMOPs. For researchers attempting to find a DMOP benchmark suite with characteristics similar to a real world problem under investigation, the baseline `difficulty' of the parameters pertaining to the dynamics and the problem are under-documented. Whilst the recently proposed comprehensive benchmark sets SDP \cite{Jiang2019}, RDP \cite{Ruan2021} and JY \cite{Jiang2017}, offer a multitude of well-designed and diverse benchmarks, the results presented examine a limited ranges of dynamic parameters in terms of severity and frequency of changes.

Many works use non-dynamic MOEAs (designed originally for static problems) to compare and contrast the performance of their proposed DMOEA \cite{Chen2017,Gee2017,Mehnen2006,HZhang2021,Jiang2019}. Some of these feature simple modifications to handle dynamic changes such as restarting or introducing random solutions. However the detriment or benefit of such modifications relative to the standard implementation of these MOEAs is unknown. Understanding the capability of MOEAs for DMOPs in a fine-grained experimentation of frequency and severity will allow us to determine the validity of prior application.


In order to determine a baseline difficulty for existing dynamic multi-objective benchmarks there are some key parameters, on which we can determine the baseline performance of state of the art static MOEAs. We argue, that it is essential a systematic examination of the dynamic problem parameters -- severity and frequency -- is carried out. Such an examination would provide insights into how specific algorithms and simple response mechanisms can cope with the changing dynamics, thus allowing for a definition of a baseline range of parameters for a given DMOPs that cannot effectively be handled by existing MOEAs. 
Here, using a subset of established DMOPs, we illustrate the importance of the frequency and severity parameters through a fine-grained experimental protocol across ranges of these parameters that encompass the commonly used values in the literature. 
There exists a framework called PlatEMO \cite{Tian2017} which gathers a comprehensive library of multi-objective optimization problems and Evolutionary Algorithms (EAs) designed to solve them. This provides the inspiration for the collation of DMOPs, and in further work, the collation of the vast range of DMOEAs currently in the literature. 
By providing the tools we employ to generate our results, we facilitate future consistency and therefore verification, corroboration and comparison of results. The proposed schema for determining the maximum capability of MOEAs on DMOPs provides the framework for determining the dynamic instances of problems that novel DMOEAs should be tested on to determine meaningful improvements over existing methods. 


To summarise the contributions of this paper:  
\begin{itemize} \vspace{-0.135cm}
    \item Firstly, we provide an experimental schema allowing for the reproducibility of results for DMOPs, including the DPTP, a platform allowing for fine-grained experimentation of DMOP parameters.
    \item Additionally, we perform for a selection of typical benchmark DMOPs, an investigation of parameters values commonly used in the literature that govern the frequency and severity of changes.
    \item We examine the performance of non-dynamic MOEAs on these parameters to ascertain a minimum recommended experimental range for frequency and severity, below which good performance can be achieved by non-dynamic MOEAs.
    \item The utility of commonly used simple modifications to static MOEAs (random and mutated solution additions) is contrasted to the random restart methodology which is widely used as baseline comparison for novel algorithm performance.
\end{itemize}



The remainder of the paper is organized as follows. Section \ref{sec:Background} contains a summary of DMOP definitions including the frequency and severity parameters, together with their usage and the usage of MOEAs on DMOPs in the literature. Section \ref{sec:Methods} gives an overview of the experimental platform and the problems, algorithms, responses and parameter ranges used. Section \ref{sec:Results} details the key results pertaining to each of the contributions listed above. Conclusions are drawn in Section \ref{sec:Conclusions}.

\section{Background} \label{sec:Background}

\subsection{Dynamic multi-objective Optimization Problems}

Dynamic multi-objective optimization problems extend multi-objective problems to include time-variant terms or components, often to incorporate behaviours of a real-world system. Equation~\ref{eqn:DMOP} gives one formulation of a bi-objective DMOP with time-dependent objective functions. 
\begin{equation}
    \begin{aligned} 
      &\vec{\textbf{x}}=[x_{1}, x_{2}, \ldots x_{n}]\\
      \vec{\textbf{F}}(\vec{\textbf{x}},t)=&[f_{1}(\vec{\textbf{x}},t), f_{2}(\vec{\textbf{x}},t), \ldots f_{M}(\vec{\textbf{x}},t)]\\
      &h_{1}(\vec{\textbf{x}})\leq0, h_{2}(\vec{\textbf{x}})=0
    \end{aligned}
    \label{eqn:DMOP}
\end{equation}
Dynamics can also present in the decision variables or constraints (e.g. $x(t)$ or $h(\vec{\textbf{x}},t)$) and in the number (e.g. $x_n{(t)}$ or $f_{M(t)}$) of any of these. 

The first DMOP benchmarks suite proposed \cite{Farina2003} included some aspects of the above possible dynamics. The FDA suite has been widely used and modified as explained by Helbig and Engelbrecht \cite{Helbig2013a}. However, in the vast literature concerning the testing of algorithms on these problems, the experimentation and understanding of the dynamics parameters is limited. 

Other dynamic multi-objective benchmarks have been proposed since, also detailed in \cite{Helbig2013a}. Complex and comprehensive suites have been recently proposed that cover a vast range of problem characteristics in terms of the POS and POF and their changes over time. Namely the SDP\cite{Jiang2019}, RDP\cite{Ruan2021} and JY\cite{Jiang2017} suites.

\subsection{Severity and Frequency}
Examination of the frequency and severity has been conducted in terms of run time and hitting time for single objective benchmarks \cite{Rohlfshagen2009}, instead here we determine the veracity of the collective adoption of a limited range of values for these parameters in multi-objective benchmarks. The time-dependency of components in DMOPs is controlled by two factors, the severity and frequency of changes. The reciprocal  of the severity, commonly denoted as $\frac{1}{n_{t}}$, controls the magnitude of changes in the value of $t$. For example, for $n_{t}=20$, the value of $t$ increases by 0.05 at each change event. Within our results we illustrate the severity as its reciprocal to more easily visualize an increasing magnitude of change. The frequency of the changes is controlled by $\tau_{t}$ and is measured in generations; a value of $\tau_{t}=20$ means that each dynamic interval (period between changes) lasts for 20 generations before the value of $t$ changes again. Whilst other parameters, such as the number of decision variables ($n$), can alter the difficulty of a problem, this challenge is not unique to DMOPs and so is not considered here.

Various recommendations are made in the literature as to the values of $n_{t}$ and $\tau_{t}$ that should be investigated. For example, Helbig \& Engelbrecht \cite{Helbig2013a} suggest values of $n_{t}=\{1,10,20\}$ and values of $\tau_{t}=\{5,10,25,50,100\}$ to be used in various combinations. The parameters combinations in these ranges enable the investigation of algorithms in environments with fast-changing and high-magnitude changes or in those with slower changes and smaller magnitude changes, where effective change detection may be a motivator. In contrast, Farina et al \cite{Farina2003} when defining the fundamental FDA suite suggest only parameter settings of $n_{t}=10, \tau_{t}=5$ with $n=20$.

Despite the recommendations, the range of severity values and frequencies are investigated are inconsistent, incomplete and sometimes unjustified. We consider two measures of experimental comprehensiveness: the number of combinations and the range of each parameter.
The number of examined $n_{t}$--$\tau_{t}$ pairings is varied in recent works including one \cite{Jiang2019,Ruan2021}, two \cite{Yazdani2020,Jiang2018}, three \cite{Zou2021,QZhang2021,Gee2017,Jiang2017} and five \cite{Jiang2014}. Only the recent work by Zhang et al. \cite{HZhang2021} considers a more diverse set with six different frequency-severity pairings: $n_{t}$–-$\tau_{t}= \{5-5, 5-10, 5-20, 10-5, 10-10, 10-20\}$. These settings are applied to a novel set of proposed ZDT-based functions with time windows rather than any existing DMOP benchmarks.

Within the available literature, the values of examined severity values are limited to $n_{t}=\{5,10,20\}$ and the most commonly evaluated frequencies are $\tau_{t}=\{5,10,20\}$. There are few papers that consider frequencies outside of this range despite the aforementioned recommendations; Chen et al. \cite{Chen2017} explores changing the number of objectives by one at four change frequencies $\tau_{t}=25,50,100,200$. Other works on single-objective problems consider pairings of frequency and severity that do not easily translate to the $n_{t}$--$\tau_{t}$ system \cite{Yazdani2020}. 
The work of Mehnen et al. \cite{Mehnen2006} considers different frequencies for each of the considered DMOP benchmarks (a variety of FDA and DTF problems) with minimum $\tau_{t}=1$ and a maximum $\tau_{t}=50$. In other cases, the severity and frequency is matched to observations from real world systems and does not easily compare with the typical severity or frequency definitions \cite{Hasan2019}.

In brief, the range of frequencies and severity parameter values used in the literature is limited and the same for all DMOP benchmark studies. The aforementioned papers use these parameter values in mixed sets of problems selected from many of the well known and recent benchmark sets including FDA \cite{Farina2003}, DIMP \cite{Koo2010}, dMOP \cite{Goh2009}, DSW \cite{Mehnen2006}, T \cite{Huang2011}, HE \cite{Helbig2013a}, ZJZ \cite{Zhou2007}, UDF \cite{Biswas2014}, SDP \cite{Jiang2019}, RDP \cite{Ruan2021}, JY \cite{Jiang2017}, DF \cite{Jiang2018} and GTA \cite{Gee2017}.

\subsection{Use of non-dynamic MOEAs for DMOPs}
Many articles from the related literature that propose novel algorithms for solving DMOPs use the performance of well-known MOEAs (and simple modifications of them) as baselines for comparison. These `static' or generic MOEAs have not been designed specifically for DMOPs. The NSGA-II \cite{Deb2002-nsgaii} has seen comparisons \cite{Mehnen2006,Chen2017} and simple modifications to include random or mutated solutions in response to changes (DNSGA-II algorithm types A and B respectively \cite{Deb2007-dnsgaii}) are also used \cite{Helbig2013a,Jiang2017,Jiang2019,HZhang2021,Chen2017,Gee2017}. The NSGA-III algorithm is also used and modified for performance comparisons \cite{HZhang2021}, as is the SPEA2 \cite{Zitzler2002} algorithm \cite{Mehnen2006,Jiang2017,Jiang2014}.
The non-dynamic MOEA/D algorithm \cite{Zhang2007} is also compared \cite{Jiang2017,Chen2017,Jiang2019} and augmented with Kalman Filter prediction \cite{Gee2017,Chen2017}, reinforcement learning \cite{Zou2021}, intensity of environmental change handling \cite{Ruan2021,Zou2021} \& a first order difference model \cite{HZhang2021}.

Widespread use of non-dynamic MOEAs in DMOP experiments is intuitive; these are the algorithms that provide the best performance on static problems and this may extend to dynamic problems too. However it is common that MOEAs are given a `random restart' response where they are forced to reinitialize in response to dynamic changes \cite{Jiang2017,Ruan2021,Gee2017,Jiang2014}. We compare the simple modification to add random and mutated solutions in response to a change, a `do nothing' approach and contrast it to the poor performance when restarting generic MOEAs for DMOPs.

We focus on the use of non-dynamic MOEAs here rather than the plethora of detailed and complex dynamic methods. There are many purpose-built methods for DMOP that include prediction mechanisms \cite{Zhou2007,Zhou2014,Koo2010,Peng2013,Hatzakis2006}, multiple archive based methods \cite{Chen2017} and ensemble methods \cite{Muruganantham2016,Azzouz2017} among others \cite{Nguyen2012,Helbig2013a,Helbig2013b,Branke2003}.

\section{Methods} \label{sec:Methods}

 \subsection{Platform specification}
Of the surveyed literature, only the works proposing the GTA \cite{Gee2017} and SDP \cite{Jiang2019} suites provided code to replicate the problem implementations. Given that the equations defining the FDA problems are inconsistent in the literature \cite{Farina2003,Helbig2013a} provision of the code used to generate the results is a simple but important way to ensure the reproducibility of experiments and results by others.

We therefore construct a framework to demonstrate MOEA performance on DMOPs. Inspired by the existing PlatEMO MATLAB implementation, our platform is adapted to include DMOPs and allow for experimentation on dynamic frequency and severity. The class structures and main running functions of the framework used here are distinct from those used in PlatEMO and we name our platform \textbf{D}MOP \textbf{P}arameter \textbf{T}esting \textbf{P}latform (DPTP). Here we document four MOEAs and four DMOPs, however the platform so far contains the  problems from the following suites: FDA \cite{Farina2003}, dMOP \cite{Goh2009}, DIMP \cite{Koo2010}, ZJZ \cite{Zhou2007} \& HE \cite{Helbig2013a}. Through the GUI, experimentation on any of the problem parameters is possible, including the number of decision variables and their ranges, the number of objectives. The selected algorithm can be augmented with any of the simple response mechanisms and additional parameters controlling the dynamics including the range of $t$, cycling behaviour and delay onset can be controlled. DPTP allows for fine-grained examination for ranges for two parameters to produce heatmaps like those presented in this paper. A variety of metrics can be recorded as in PlatEMO, including GD, IGD, HV and others. The well-designed and comprehensive JY \cite{Jiang2017}, RDP \cite{Ruan2021}, GTA \cite{Gee2017} and SDP \cite{Jiang2019} benchmark sets will be added in future in addition to a number of dynamic algorithms. The DPTP is available at \hyperlink{https://github.com/DPTP2022/DPTP}{https://github.com/DPTP2022/DPTP}.

  \begin{table*}[!t]
     \centering
     \caption{DMOP benchmark problems. Based on literature values for all problems, the number of objectives is limited to 2 and the number of decision variables is fixed at 20.} \vspace{-0.3cm}
     \begin{tabular}{|c|c|c|c|c|}
        \hline
        Name    & Objective Functions   & Decision Variables & Type  & Reference   \\ \hline
        dMOP1   & \begin{tabular}{@{}c@{}}$ f_{1}(\mathbf{x_{I}})=x_{1}; \quad f_{2}=gh$ \\ $g(\mathbf{x_{II}})=1+9\sum{}_{x_{i}\in\mathbf{x_{II}}}(x_{i})^{2};$ \quad  $h=1-\left(\frac{f_{1}}{g}\right)^{H(t)}$ \\ $H(t)=0.75sin(0.5t)+1.25$ \end{tabular}                       & \begin{tabular}{@{}c@{}} $x_{i}\in[0,1]$ \\ $\mathbf{x_{I}}=(x_{1}) ; \quad \mathbf{x_{II}}=(x_{2},\ldots,x_{n})$ \end{tabular}    & III  & \cite{Goh2009}  \\ \hline
        dMOP2   & \begin{tabular}{@{}c@{}}$ f_{1}(\mathbf{x_{I}})=x_{1}; \quad f_{2}=gh$ \\ $g(\mathbf{x_{II}})=1+9\sum{}_{x_{i}\in\mathbf{x_{II}}}(x_{i}-G(t))^{2};$ \quad  $h=1-\left(\frac{f_{1}}{g}\right)^{H(t)}$ \\ $H(t)=0.75sin(0.5t)+1.25$; \quad $G(t)=sin(0.5\pi t)$\end{tabular} & \begin{tabular}{@{}c@{}} $x_{i}\in[0,1]$ \\ $\mathbf{x_{I}}=(x_{1}) ; \quad \mathbf{x_{II}}=(x_{2},\ldots,x_{n})$ \end{tabular}      & II  & \cite{Goh2009}  \\ \hline
        DIMP2   & \begin{tabular}{@{}c@{}}$ f_{1}(\mathbf{x_{I}})=x_{1}; \quad f_{2}=gh$ \\ $g(\mathbf{x_{II}})=1+2(n-1)+\sum{}_{x_{i}\in\mathbf{x_{II}}}[(x_{i}-G(t))^{2}-2cos(3\pi(x_{i}-G_{i}(t)))]$   \\ $h=1-\sqrt{\frac{f_{1}}{g}};$ \quad $G_{i}(t)=sin(0.5\pi t+2\pi\frac{i}{n+1})^{2}$\end{tabular} & \begin{tabular}{@{}c@{}} $x_{I}\in[0,1]$; \quad $x_{II}\in[-2,2]$  \\ $\mathbf{x_{I}}=(x_{1}) ; \quad \mathbf{x_{II}}=(x_{2},\ldots,x_{n})$ \end{tabular}      & I  & \cite{Koo2010}  \\ \hline
        HE1     & \begin{tabular}{@{}c@{}}$ f_{1}(\mathbf{x_{I}})=x_{1}; \quad f_{2}=gh$ \\ $g(\mathbf{x_{II}})=1+\frac{9}{n-1}\sum{}_{x_{i}\in\mathbf{x_{II}}}x_{i};$ \quad  $h=1-\sqrt{\frac{f_{1}}{g}}-\frac{f_{1}}{g}sin(10\pi tf_{1})$ \end{tabular} & \begin{tabular}{@{}c@{}} $x_{i}\in[0,1]$ \\ $\mathbf{x_{I}}=(x_{1}) ; \quad \mathbf{x_{II}}=(x_{2},\ldots,x_{n})$ \end{tabular}      & III  & \cite{Helbig2013a}  \\ \hline
     \end{tabular}
      \vspace{-0.3cm}
     \label{tab:problems}
 \end{table*}

 \subsection{Problems, Algorithms \& Measurements} \label{sec:Algorithms}
 As recent DMOP benchmark suites such as the SDP, RDP and JY sets have not yet been used in many works, we instead choose to provide insights on a subset of the commonly used benchmarks. Further works will extend this methodology to as much of the DMOP space as possible.

 The problems featured in these experiments are given in Table \ref{tab:problems}, together with their objective functions, type classification and references. The number of decision variables is fixed to 20 for these problems as this is the value almost unanimously used in literature studying DMOPs. 
 
 Four well-known algorithms designed for non-dynamic MOOPs are employed here. The first is NSGA-II \cite{Deb2002-nsgaii}, a population based algorithm that uses non-dominated sorting Pareto ranking based selection from a population doubled by crossover and mutation operators, and replacement based on crowding distance. Secondly, NSGA-III \cite{Deb2014-nsgaiii}, another Pareto-dominance based algorithm, improves over the previous algorithm by replacing crowding distance based replacement with a distributed and maintained set of reference points. Thirdly, MOEA/D is a decomposition based algorithm, which uses a series of reference vectors to guide search towards the Pareto Set. Finally, SPEA2 is another Pareto-dominance based algorithm that maintains an external archive of solutions in addition to an exploratory population of solutions. In addition to this generic dynamic response mechanisms are incorporated for comparison to the default algorithms. These are based on the commonly used diversity introduction mechanisms as utilised by the DNSGA-II algorithm; replacement of solutions with random or mutated solutions. A random restart (re-initialization of the population) method is also commonly used in the literature and therefore is included too. A population size of 100 is the most commonly used across the surveyed literature and we adopt this value also.
 

 In terms of performance measurement, a variety of methods have been documented \cite{Camara2009,Helbig2013b} and these are largely used as summary statistics for DMOEAs. As the motivation of this work is to ascertain the minimum DMOPs parameters that require dedicated algorithm design beyond the capabilities of existing MOEAs, we require only simple measurements. Whilst the proposed platform can record a range of measurements, we report results limited to the commonly used Mean Hypervolume Difference, detailed in Eq. \ref{eqn:HypervolumeDifference}. We take the difference of the hypervolume and the optimal hypervolume (using the same number of solutions) at the generation before a dynamic change, representing the minimum error to the best hypervolume achieved in each time interval. As we investigate MOEA performance rather than mechanisms to detect dynamic changes \cite{Morrison2013-detection}, dynamic responses are prompted for these experiments.

 \begin{equation}
     HVD_{n} = (Optimal Hypervolume)_{n}-(Achieved Hypervolume)_{n}
     \label{eqn:HypervolumeDifference}
 \end{equation}
 
 where $n$ is the number of solutions used in the optimal calculation and as the population size for the algorithms. The optimal value for this is 0, meaning that the solutions found by the algorithm perfectly match the optimally distributed Pareto Front sample. The magnitude of any deviation from true hypervolume is specific to the problem and so we forego normalization in this instance.


 \subsection{Dynamic Changes}
 
 Our experiments focus on determining the limitations of MOEAs for DMOPs according to the commonly used problem parameters in the literature. Therefore we investigate effects of different frequencies and severity values of changes on the benchmark DMOPs. As detailed, to cover the commonly used values in the literature; 21 values of $1/n_{t} \in [0.01 0.5]$ and 30 values of $\tau_{t} \in [1,30]$. This means the most rapid change frequency is every generation ($\tau_{t}=1$) and the least rapid every 30 generations. As mentioned, some have suggested $\tau_{t}=50$ \cite{Chen2017,Mehnen2006}, however for data collection feasibility we limit our investigation to commonly used values.

 \subsection{Experimental plan}

 Using our streamlined experimental DPTP platform, we gather the HVD data for a number of DMOP benchmarks, the results presented here correspond to the problems in Table \ref{tab:problems}. For each problem we examine every pairwise combination of change frequency and severity parameters in the examined ranges detailed above. For each combination of change frequency and severity each of the four algorithms detailed in \ref{sec:Algorithms} with each of the four baseline response mechanisms (DR0: none, DR1: random solution addition, DR2: mutated solution addition and DR3: random restart) is applied to the problem. This results in 21x30x4x4=10,080 algorithm runs per repeat, per problem, which we summarise in a series of heatmaps for clarity.

\begin{figure*}[!t]
\centering
\includegraphics[width=0.95\textwidth]{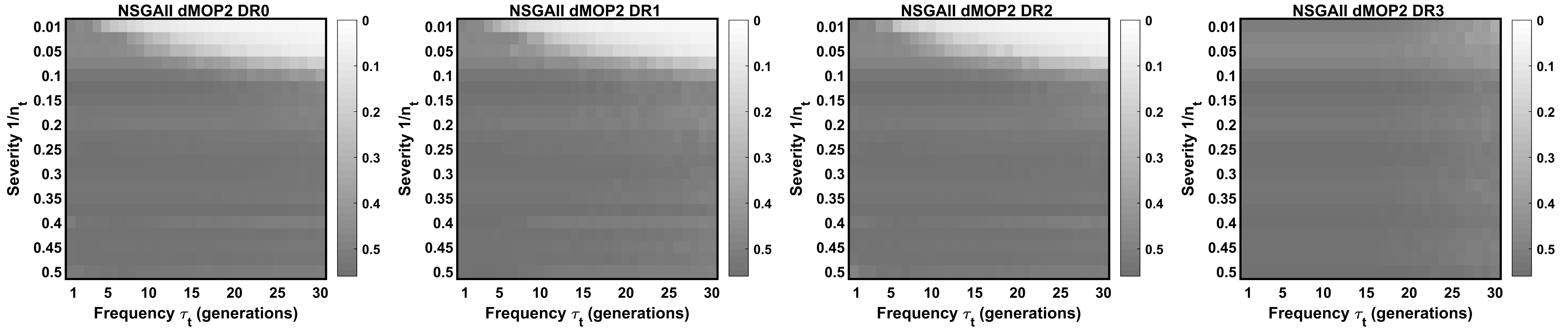}
\caption{Frequency and severity heatmaps of HVD on dMOP2 for each response type (DR0: none, DR1: random solution addition, DR2: mutated solution addition, DR3: random restart). A zero-value shows optimal hypervolume attainment and higher values show poorer performance.}
\label{fig:ResponseComparison_dMOP2_NSGAII}
\end{figure*}

\begin{figure*}[!t]
\centering
\includegraphics[width=0.95\textwidth]{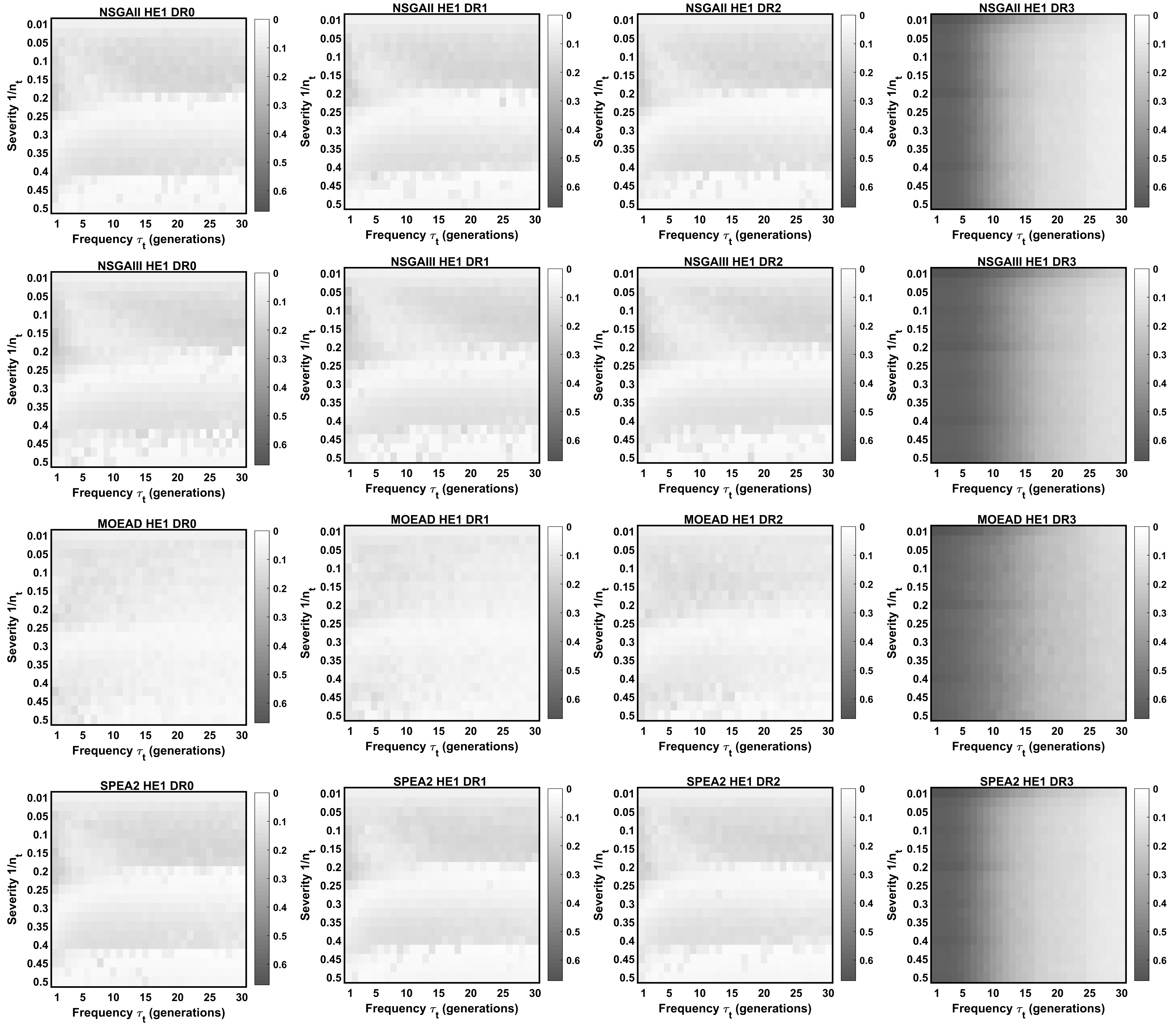}
\caption{Frequency and severity heatmaps of HVD on HE1 for each algorithm and each response type (DR0: none, DR1: random solution addition, DR2: mutated solution addition, DR3: random restart). A zero-value shows optimal hypervolume attainment, higher values show poorer performance.}
\label{fig:AlgorithmAndResponseComparison_HE1}
\end{figure*}

\section{Results} \label{sec:Results}
    
Given that a limited range of parameters for the change severity and change frequency are investigated in the literature, the following results highlight the difficulty of the most popularly used change severity and frequency instances in terms of their ability to be solved by generic MOEAs.

We recognise that problems have different difficulty in spite of dynamics and by showing only select problems we narrow our conclusions. However, we aim to determine a sensible range of dynamic parameters for the presented DMOP benchmarks and provide a methodology such that the performance of newly devised algorithms is tested in a meaningful way. Moreover, we provide insights on baseline algorithm comparison and selecting dynamics parameters that novel algorithms should be tested on to show a meaningful development and contribution to the field.

\begin{figure}[!t]
\includegraphics[width=0.48\textwidth]{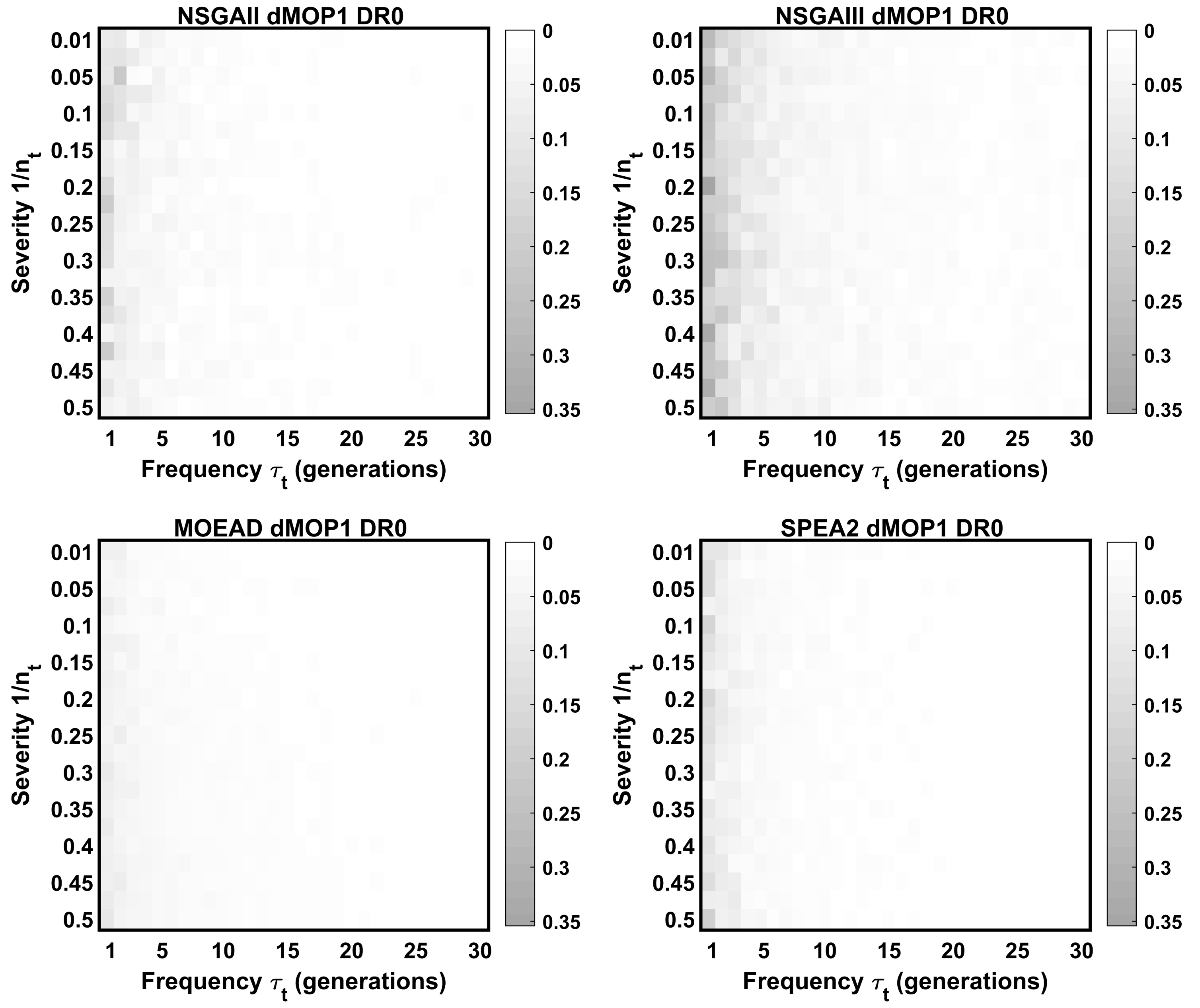}
\caption{Heatmaps of HVD on the dMOP1 benchmark for each algorithm with no dynamic response mechanism. A zero-value shows complete attainment of optimal hypervolume. Combinations of frequency (x-axis) and severity (y-axis) are examined in ranges that include the commonly used values in the literature. Each cell represents the mean of 30x5 change events (150 pre-change-event measurements).}
\label{fig:AlgorithmComparison_dMOP1_DR0}
\end{figure}

\begin{figure}[!t]
\includegraphics[width=0.47\textwidth]{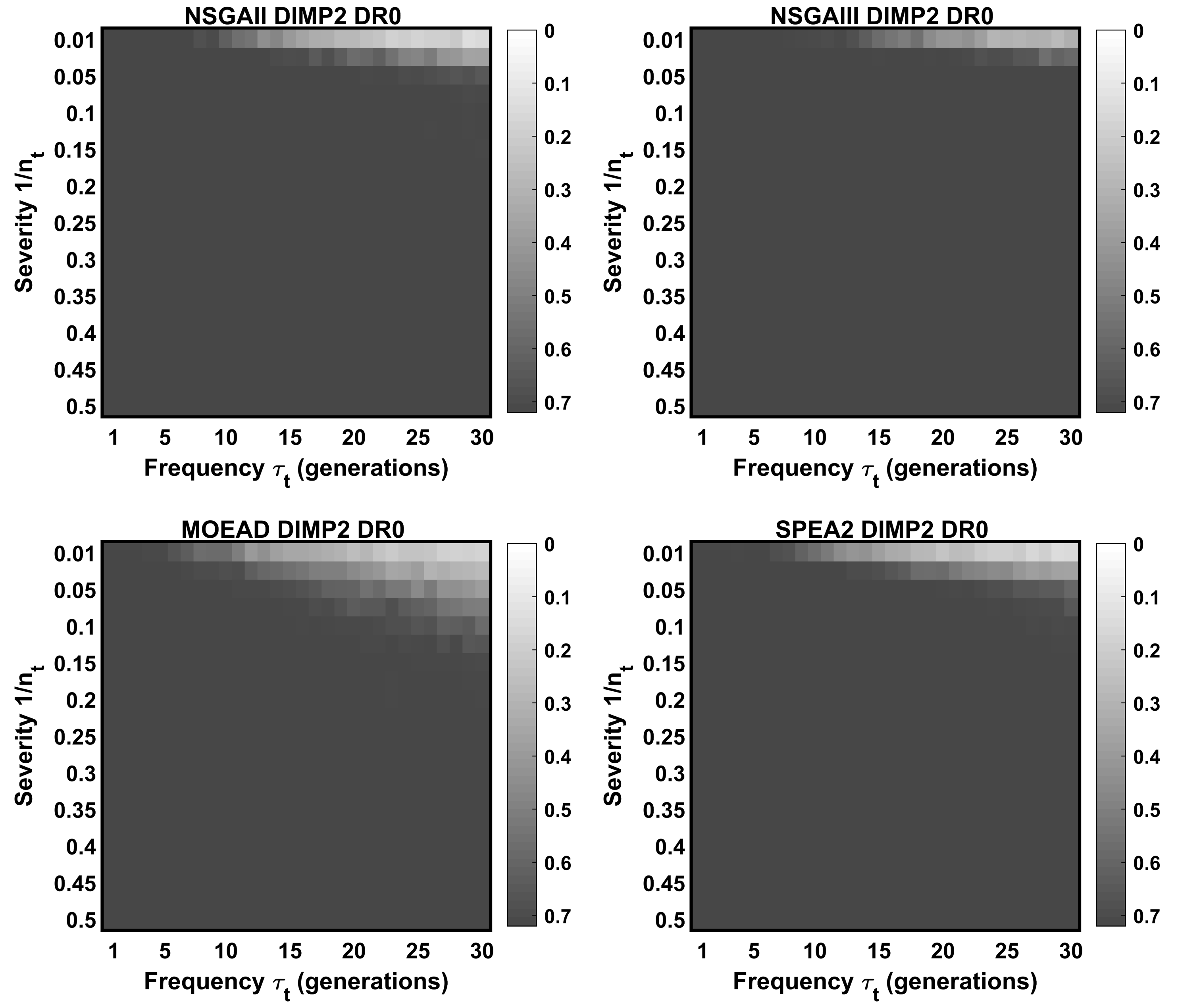}
\caption{Heatmaps of HVD for the DIMP2 benchmark for each algorithm with no dynamic response mechanism. A zero-value shows complete attainment of optimal hypervolume. Combinations of frequency (x-axis) and severity (y-axis) are examined in ranges that include the commonly used values in the literature. Each cell represents the mean of 30x5 change events (150 pre-change-event measurements).}
\label{fig:AlgorithmComparison_DIMP2_DR0}
\end{figure}

\begin{figure*}[!t]
\includegraphics[width=\textwidth]{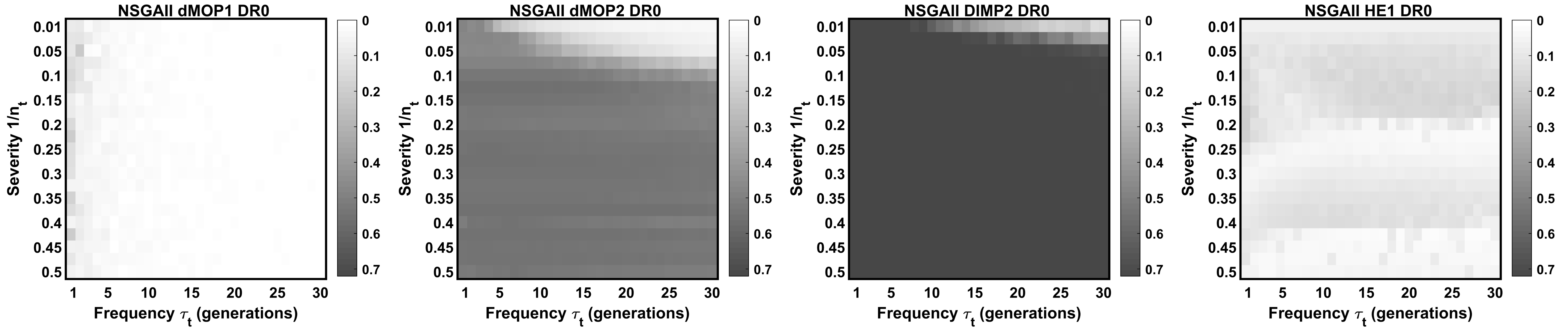}
\caption{Frequency and severity heatmaps of HVD for the NSGA-II algorithm with no dynamic response mechanism on problems with Type I(DIMP2), II (dMOP2) and III (dMOP1/HE1) dynamics. A zero-value shows complete attainment of optimal hypervolume, higher values indicated poorer performance. Each cell represents the mean of 30x5 change events (150 pre-change-event measurements).}
\label{fig:ProblemComparison_NSGAII_DR0}
\end{figure*}

\subsection{Justifying baseline dynamic responses for comparison}

The `random restart' mechanism is used in conjunction with a simple MOEA as a baseline for comparison in many papers that propose novel dynamic algorithms, however we demonstrate here through the hypervolume attainment, that it is not useful to use such a method for comparison. Figure \ref{fig:ResponseComparison_dMOP2_NSGAII}, illustrates that of the considered simple baseline responses investigated here, restarting the algorithm (DR3 subplot) is clearly the worst. The hypervolume attainment is poor across the range of change severity values and frequencies compared with no response (DR0), and the random (DR1) and mutated (DR2) solution responses. Furthermore, little improvement is seen by purely addition of 20\% mutated or randomised solutions; the DIMP2 problem appears difficult with literature-standard dynamics parameters for non-dynamic MOEAs.
Figure \ref{fig:AlgorithmAndResponseComparison_HE1} illustrates the performance of each response type in each algorithm for the HE1 problem. There is a more complex pattern of hypervolume attainment in the parameter space, with two bands of better performance at high severity ($\frac{1}{n_{t}}=0.5$) and mid-severity ($\frac{1}{n_{t}}=0.25$), but immediately worse performance vertically adjacent to these regions at ($\frac{1}{n_{t}}=0.4$) and ($\frac{1}{n_{t}}=0.2$). These patterns are consistent across the DR0, DR1 \& DR2 response for the NSGA-II, NSGA-III \& SPEA2 algorithms. 
The performance of MOEA/D algorithm is more homogeneous across the parameter combinations with these patterns less clearly defined. One possibility for this is the drastic and non-uniform change in POF shape the problem experiences with changing values of $t$. The non-uniform performance patterning means that for some values of the severity parameter, these specific successive changes in the problem's POF are more easily handled. 
This implies that the $n_{t}$ parameter does not, for all problems, correspond to a series of change events of equal magnitude in terms of their difficulty for algorithms. This reinforces the need for careful selection of severity and frequency parameter values when benchmarking algorithms on DMOPs. 
All algorithms performance with the random restart response are also shown in the right most column of the grid in Figure \ref{fig:AlgorithmAndResponseComparison_HE1}. The negative impacts on algorithm performance are less severe at lower change frequencies ($\tau_{t}=30$), however the hypervolume difference remains high when compared with the other response types, as in Figure \ref{fig:ResponseComparison_dMOP2_NSGAII}. This strengthens the evidence against using random restart in algorithms as a baseline for performance comparison.

\subsection{Determining baseline parameters for dynamics in DMOPs}

Figure \ref{fig:AlgorithmComparison_dMOP1_DR0} illustrates a key motivation of this work in providing insights into the capability of MOEA performance on DMOPs with commonly used values for the frequency and severity of changes. The dMOP1 problem poses little difficulty at any of the severity or frequency settings, with the exception of a change frequency of 1, where it can be expected that a generic algorithm might lose performance. Change severity had little impact, whilst smaller frequencies (more rapid changes) somewhat reduced the hypervolume attainment of the algorithms. Notably, this performance degradation occurs sooner for NSGA-III than for NSGA-II or the other algorithms. Our recommendation for dMOP1 is that most of the studied dynamics ranges in the literature can be effectively handled by generic MEOAs and therefore should not be used to test more complex novel dynamic algorithms.

Conversely, the DIMP2 problem poses a challenge to all the algorithms across most of the severity and frequency ranges. Figure \ref{fig:AlgorithmComparison_DIMP2_DR0} illustrates that for all algorithms, it is only for small severity and the least rapid changes that minimal HVD is possible, with MOEA/D having the best performance area. For the majority of the parameter ranges these generic DMOEAs cannot provide good results. Therefore our recommendation is that DIMP2 is suitable for benchmarking novel dynamic methods.

\subsection{Algorithm performance on different dynamics types}
We illustrate the variable performance of the NSGA-II algorithm on the problem subset in Figure \ref{fig:ProblemComparison_NSGAII_DR0}. The difference in performance of a single algorithm on multiple problems is to be expected, however, more importantly these heatmaps illustrate the impact of dynamic parameter choices. Using the same values of change severity and frequency for different problems without justification can weaken conclusions on algorithm performance. For example, the NSGA-II algorithm, not designed for DMOPs, provides competitive performance on dMOP1 across all but the lowest frequency changes. Comparison of a novel DMOEA on this problem is effectively meaningless, other than to ensure a basic capability. Other existing DMOP benchmarks however, pose significant challenges for the examined MOEAs. For example, NSGA-II struggles on DIMP2 for most combinations of $n_{t}$ and $\tau_{t}$. It is these problems that should therefore be the focus when testing the efficacy of novel DMOEA constructions.

Furthermore, our results stress the importance of both selecting the parameters of the frequency and severity of changes and determining their importance in a problem's difficulty. Concluding the effectiveness of a novel dynamic method on, for example, the dMOP2 problem with $n_{t}=20$ and $\tau_{t}=25$ is misleading. This combination of dynamic frequency and severity can effectively be handled by a generic MOEA.

\subsection{Recommendations for Future Investigations}
For each of the examined problems, we provide a summary of the minimum parameters we recommend to ensure a challenging problem set when evaluating algorithms.

\textbf{dMOP1 -} This problem did not provide a challenge for any of the MOEAs tested. A drop off in performance occurs when the change frequency $\tau_{t}=1$, however this would also be challenging for most DMOEAs except those specifically designed for problems with rapid changes. Our recommendation is therefore to avoid this problem except for ensuring fundamental capability compared to the MOEAs.

\textbf{dMOP2 -} The results on this problem indicate it is a good candidate for evaluating DMOEAs since the majority of the parameter space cannot effectively be handled by the MOEAs. Reciprocal severity values of at least $\frac{1}{n_{t}}=0.05$ are recommended in conjunction with most change frequencies. If $\tau_{t}$ is less than 15, smaller severity values can be considered.

\textbf{DIMP2 -} This problem was the most challenging for the MOEAs of the problems examined. Using $\frac{1}{n_{t}}=0.01$ was challenging at frequencies below 20 generations, and at all frequencies for greater severity values. The MOEA/D algorithm achieved slightly better performance at low frequencies (slower changes) and low severity values, however the overall HVD is poor. 

\textbf{HE1 -} The results indicate a complex range of performance across the frequency and severity parameter space, perhaps due to the complex PF shape this problem has and the change that happens with different intervals of $t$. Careful selection of parameters may allow challenging instances, however the HV attainment is good for the MOEAs generally and therefore other problems may be better to evaluate DMOEAs.

We have evaluated the simplest modifications to MOEAs that are often used as baseline comparisons when evaluating novel DMOEAs. The results show that the commonly used `random restart' response, where the entire population is re-initialized at every change event provides poor performance across the examined parameter space. In terms of function evaluations, an equivalent number would be used in a `do nothing' approach, the DR0 response here. We therefore recommend that the random restart responses should not be used in conjunction with the examined parameter ranges. If there is justification that re-initialization may be beneficial, for example on problems with deceptive, multi-modal or many local optima, it should not be the only baseline considered - a passive approach may provide better results in many cases.

For the examined algorithms, the passive approach (DR0), the mutated solution addition (DR1), random solution addition (DR2) have relatively similar impacts on the HVD measurements, even on the HE1 problem which had a more complex performance pattern across the dynamic parameter space (as in Figure \ref{fig:AlgorithmAndResponseComparison_HE1})

Ultimately, we echo the statements of many when we conclude that a diverse set of benchmarks should be used to examine algorithm performance \cite{Helbig2013a,Raquel2013,Nguyen2012,Azzouz2016}. However we build on this to include that the parameters governing the frequency and severity of changes must be carefully selected such that the resulting instances are sufficiently challenging and cannot be handled by static MOEAs. 

\section{Conclusion} \label{sec:Conclusions}

Within this work, we have demonstrated the range of change frequency and severity parameter usage in the literature is unguided and for some dynamic multi-objective benchmarks does not result in challenging instances. Understanding the impacts of the parameters that control the dynamics in a set of selected benchmarks used to evaluate novel algorithms should be an important consideration in future works. 
Secondly we have evaluated the maximal performance of typical MOEAs not designed for DMOPs within the frequency and severity parameter range to provide an experimental schema to find recommendations for challenging instances that novel DMOEAs should consider in order to provide meaningful improvement. 
Our results elucidate the utility of different commonly used simple modifications to MOEAs in the literature which are often used as a baseline comparison for a proposed algorithm. We highlight the minimal impact on the considered subset of problems that mutated and random solutions provide and condemn the usage of the random restart response without valid justification.
Finally, we share the DPTP with which we produced these results and declare the intention to expand upon its capability in terms of the scope of DMOPs and dynamic algorithms. Our motivation is to make experimentation on DMOPs more streamlined, reproducible and verifiable to facilitate comparison such that progress towards more effective algorithms and more challenging problems can occur.

\bibliographystyle{ACM-Reference-Format}
\bibliography{refs}  

\end{document}